\pdfoutput=1

\documentclass[11pt]{article}

\usepackage[final]{acl}

\usepackage{times}
\usepackage{latexsym}
\usepackage{float}
\usepackage[T1]{fontenc}

\usepackage[utf8]{inputenc}

\usepackage{microtype}
\usepackage{caption}  
\usepackage{booktabs}  
\usepackage{float}
\usepackage{inconsolata}

\usepackage{graphicx}
\usepackage{multirow}  
\usepackage{enumitem}
\usepackage[utf8]{inputenc}
\usepackage[most]{tcolorbox}
\usepackage{titlesec}
\usepackage{xcolor}
\tcbset{
  promptbox/.style 2 args={
    enhanced,
    colback=#1!10,
    colframe=#1!80!black,
    coltitle=white,
    fonttitle=\bfseries,
    title={#2},
    sharp corners=south,
    boxrule=0.7mm,
    width=\textwidth,
    left=1mm,
    right=1mm,
    top=1mm,
    bottom=1mm,
    boxsep=4pt
  }
}

\title{CLAIM: An Intent-Driven Multi-Agent Framework for Analyzing Manipulation in Courtroom Dialogues}
\author{
Disha Sheshanarayana\thanks{These authors contributed equally to this work},\ 
Tanishka Magar\footnotemark[1],\ 
Ayushi Mittal,\ 
Neelam Chaplot \\
Manipal University Jaipur, India \\
\texttt{disha.229301161@muj.manipal.edu, tanishka.229301736@muj.manipal.edu,} \\
\texttt{ayushi.229209033@muj.manipal.edu, neelam.chaplot@jaipur.manipal.edu}
}

\begin{document}
\maketitle
\begin{abstract}
Courtrooms are places where lives are determined and fates are sealed, yet they are not impervious to manipulation. Strategic use of manipulation in legal jargon can sway the opinions of judges and affect the decisions. Despite the growing advancements in NLP, its application in detecting and analyzing manipulation within the legal domain remains largely unexplored. Our work addresses this gap by introducing LegalCon, a dataset of 1,063 annotated courtroom conversations labeled for manipulation detection, identification of primary manipulators, and classification of manipulative techniques, with a focus on long conversations.
Furthermore, we propose CLAIM, a two-stage, Intent-driven Multi-agent framework designed to enhance manipulation analysis by enabling context-aware and informed decision-making. Our results highlight the potential of incorporating agentic frameworks to improve fairness and transparency in judicial processes. We hope that this contributes to the broader application of NLP in legal discourse analysis and the development of robust tools to support fairness in legal decision-making. Our code and data are available at \href{https://github.com/Disha1001/CLAIM}{CLAIM}.
\end{abstract}

\section{Introduction}
Courtroom decisions have significant legal and societal implications, shaping legal precedents and affecting lives. However, the inherently adversarial and strategic nature of legal discourse fosters an environment where linguistic manipulation is prevalent. Tactical orchestration of manipulation can shape perceptions, steer arguments, and ultimately influence judicial outcomes.  
Over the years, studies like \cite{gold1987psychology}, \cite{lively2020seeking} and \cite{wood2012courtroom} have explored the various techniques, covert and overt, employed to manipulate courtroom dynamics, which can manifest through crafted narratives and psychological attacks. \cite{vinson1982juries} presented that defense tactics, such as contextual stimuli, can be used by lawyers to psychologically influence jurors, making it difficult for them to be unbiased or nonaligned.

\begin{figure}[t]
    \centering
    \includegraphics[width=0.95\columnwidth]{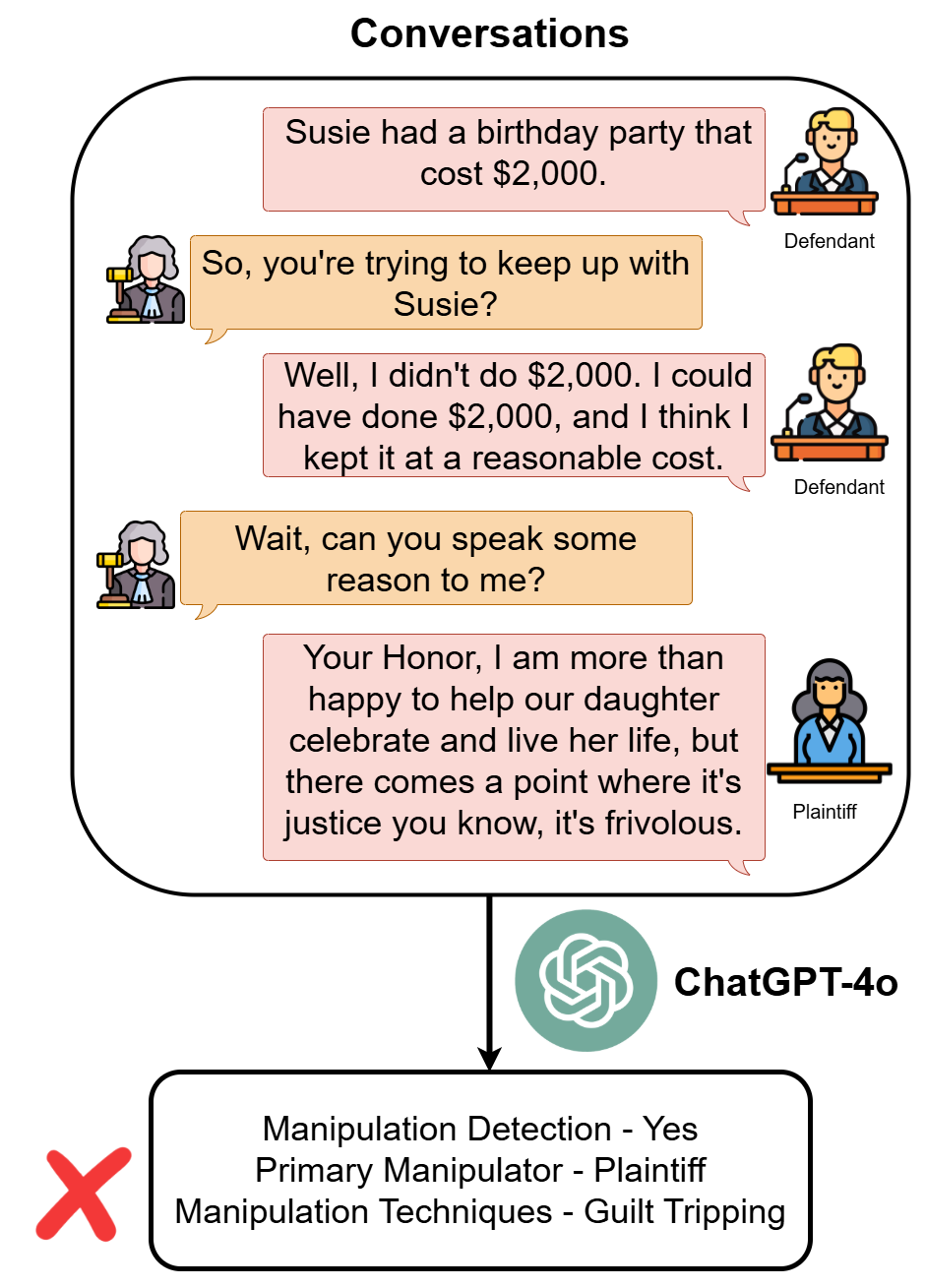}
    \caption{An example of a courtroom conversation that contains manipulation, but ChatGPT-4o fails to identify the primary manipulator and technique accurately.}
    \label{fig:SoTA}
\end{figure}

\begin{table*}[htbp]
    \centering
    \begin{tabular}{p{5cm} p{3cm} c c c} 
        \hline
        \textbf{Paper} & \textbf{Dataset} & \textbf{Detection} & \textbf{Manipulator} & \textbf{Technique} \\ 
        \hline 
        MentalManip \cite{wang2024mentalmanip} & MentalManip & Yes & No & Yes \\ \hline
        Intent-Aware Prompting \cite{ma2024detecting}  & MentalManip & Yes & No & No \\ \hline
        Advanced Prompting \cite{yang2024enhanced} & MentalManip & Yes & No & No \\ \hline
        Communication is All You Need \cite{ma2025communication}  & Multi LLM & Yes & No & Yes \\ \hline
        Human Decision-Making and AI \cite{sabour2025human}  & Custom & Yes & No & No \\ \hline
        MANITWEET \cite{huang2023manitweet} & MANITWEET & Yes & No & Yes \\ \hline
        CLAIM (Our Work) & LegalCon & Yes & Yes & Yes \\ \hline
    \end{tabular}
    \caption{Comparison of related work on manipulation analysis.}
    \label{tab:related works}
\end{table*}

\cite{gold1987psychology} emphasized that while measures against these tactics such as judicial training in psychology, court-appointed experts, increased jury compensation, and expanded jury panels may be costly, the consequences of flawed jury decision-making can be just as significant. 
Despite its serious implications for justice, computational approaches for detecting and analyzing manipulation tactics in courtrooms remain significantly underdeveloped.

While multiple studies have explored social manipulation—including fake news detection \cite{zhang2024toward}, toxic language identification \cite{li2024large} as well as the detection and categorization of mental manipulation techniques \cite{wang2024mentalmanip}—these efforts rarely focus on the legal domain. The complexity of legal language means that manipulation can be concealed behind legal jargon and thus is even more challenging to detect. 
Current SoTA models struggle to detect manipulation in courtroom debates, particularly in longer conversations, and often fail to capture the nuanced, context-dependent nature of courtroom discourse, as demonstrated in \hyperref[fig:SoTA]{Figure~\ref{fig:SoTA}}.

Our study aims to contribute to this research gap by analyzing manipulation in courtroom conversations, with a focus on long and comprehensive exchanges. We introduce \textbf{LegalCon}flict, a dataset consisting of conversations and debates sourced from transcripts across various judicial settings. It comprises 1,063 conversations annotated for manipulation detection, identification of the primary manipulator, and classification of manipulation techniques.
To evaluate this dataset, we conducted extensive experiments using SoTA models. However, these models struggled to accurately identify manipulation, particularly in complex and context-dependent cases, highlighting the need for a more specialized approach. To address this challenge, we propose \textbf{CLAIM} (Courtroom Language Analysis with Intent-driven Multi-agent Framework), a novel two-stage framework that combines an Intent-Driven Chain-of-thought prompting \cite{wei2022chain} with a Multi-Agent framework to provide a comprehensive analysis of manipulation in courtroom dialogues.
Our methodology first processes courtroom transcripts through an Intent-driven and CoT prompting technique, generating preliminary manipulation assessments. These are then passed to the Multi-Agent framework for refinement and evidence gathering. This sequential approach allows for increasingly sophisticated analysis by combining the strengths of intent-specific prompting with the collaborative reasoning capabilities of multiple specialized agents. Experimental results across diverse legal contexts show that our approach achieves significant improvements over baseline methods, particularly in detecting the primary manipulator. 

\section{Related Works}

Detecting and analyzing manipulation in conversations has been an emerging research focus, especially with the rise of large language models (LLMs) and their role in social, legal, and media contexts. Several datasets and frameworks have emerged to explore different kinds of manipulation like mental manipulation and, persuasion, misinformation, and toxicity. \hyperref[tab:related works]{Table~\ref{tab:related works}} summarizes some existing work on manipulation.
\cite{wang2024mentalmanip} introduced MentalManip, a benchmark dataset that enables fine-grained classification of manipulation in conversation.  This study focuses on identifying various manipulation techniques and vulnerabilities used in conversation, providing a solid foundation for manipulation detection in a conversation. Building on this, \cite{ma2024detecting} proposed an Intent-Aware Prompting approach that leverages speaker intent for improved detection, while \cite{yang2024enhanced} demonstrated the effectiveness of Chain-of-Thought (CoT) prompting for nuanced understanding of manipulation in conversations. However, these approaches do not attempt to identify the primary manipulator or extract manipulative techniques.
The study \cite{ma2025communication} presents a multi-LLM framework for generating persuasive dialogues. It includes both persuasion detection and technique identification, demonstrating the utility of collaborative LLM setups. 

The increasing prevalence of misinformation has also led to the development of specialized models designed for social manipulation and fake news detection, \cite{zhang2024toward} outlines strategies for mitigating manipulation in the LLM era, emphasizing the need for explainability and interpretability. \cite{huang2023manitweet} introduces a benchmark dataset MANITWEET for detecting manipulative tweets based on their distortion of news articles, highlighting the limitations of fact-checking systems and the need for better manipulation detection in social contexts.

Agent-based approaches have also shown promise. \cite{li2024large} leverages LLM agents for fake news detection, while \cite{jeptoo2024enhancing} proposes a multi-agent debate framework, where agents critique each other's outputs to improve factual accuracy. 
While these studies provide important insights into manipulation across domains, none of them focus on courtroom dialogues.

\begin{figure*}[t]
    \centering
    \includegraphics[width=\textwidth]{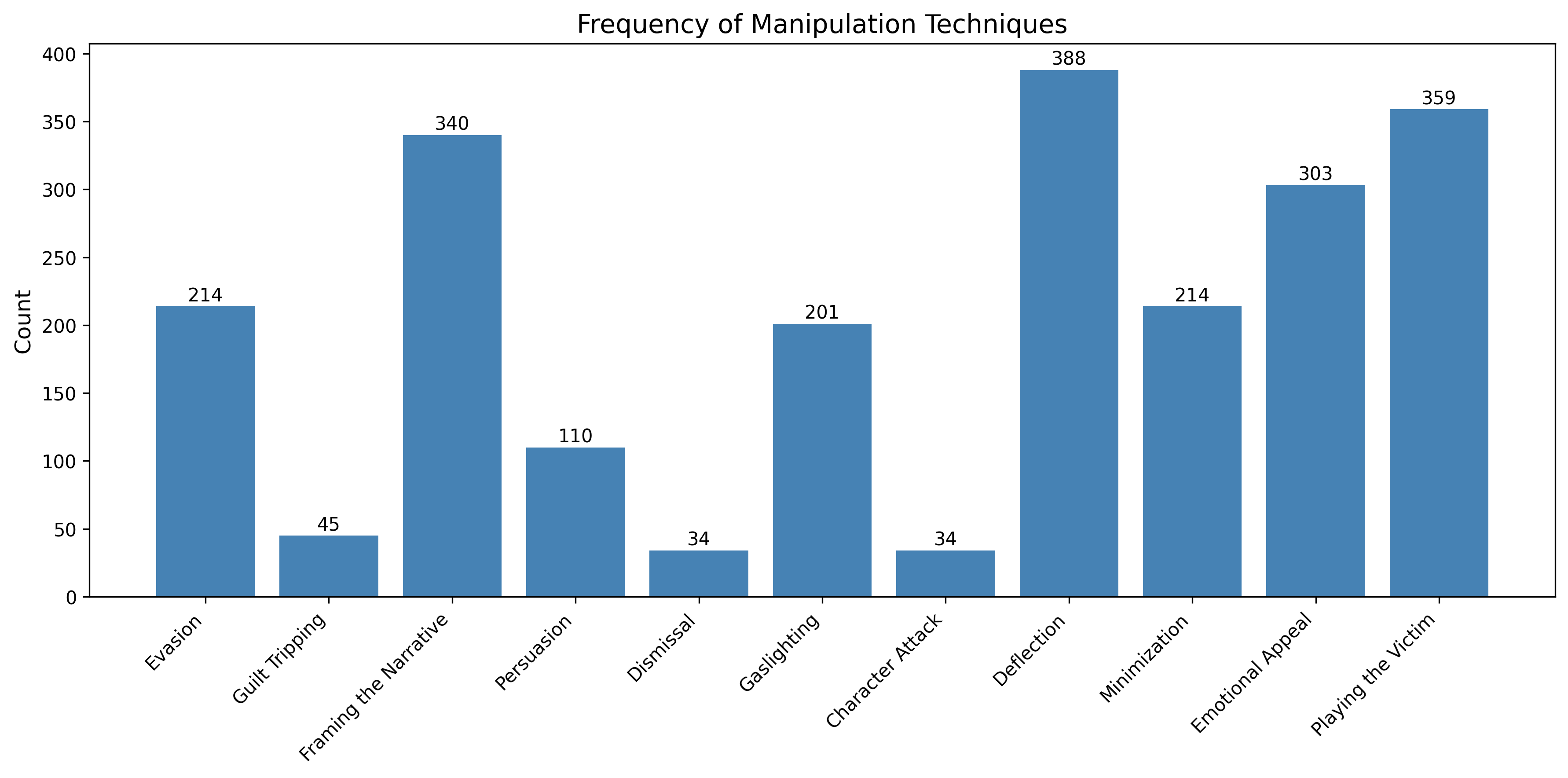}
    \caption{Bar graph showing frequency of different Manipulative Techniques in LegalCon dataset.}
    \label{fig:bar}
\end{figure*}

\begin{figure*}[t]
  \includegraphics[height=6cm]{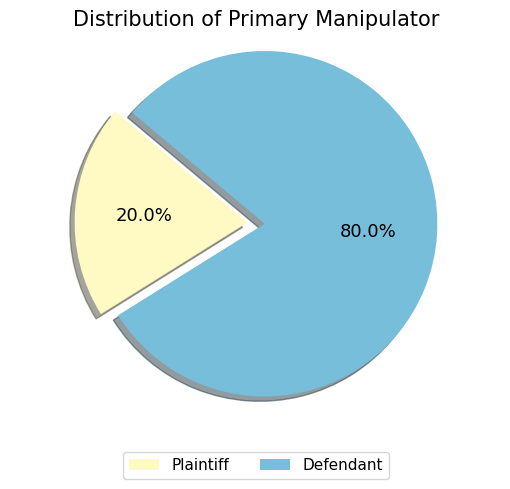} \hfill
  \includegraphics[height=6cm]{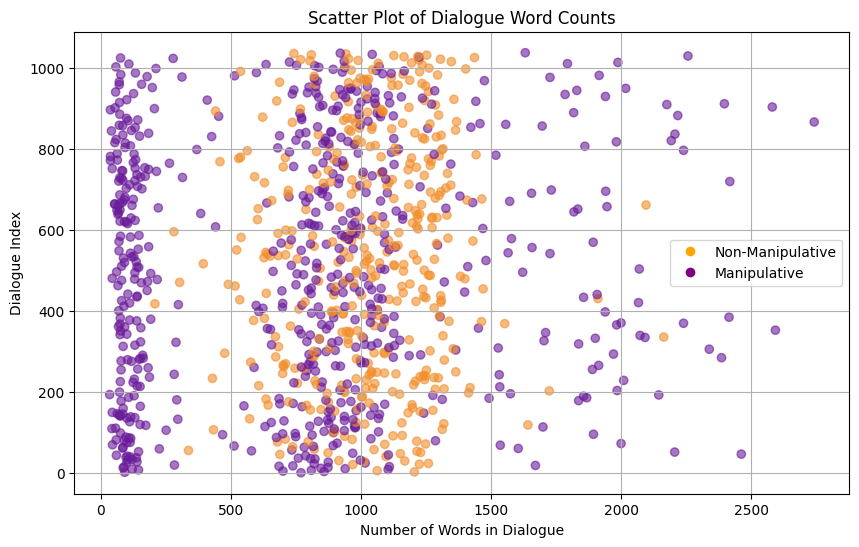}
  \caption{Pie Chart of Primary Manipulator distribution and Scatter Plot of Words Counts in Dialogues in LegalCon.}
  \label{fig:graphs}
\end{figure*}

\section{Constructing LegalCon}
\label{sec:Constructing LegalCon}
\subsection{Data Sourcing and Pre-processing}
The dataset was curated from multiple public sources and includes transcripts from various courts across the United States, such as Supreme Courts, Family Support Courts, Trial Courts, and Small Claims Courts. This selection was made to include multiple judicial contexts and case types, ensuring a comprehensive view of courtroom discourse. Long-form courtroom conversations were prioritized to capture in-depth arguments and interactions. All the transcripts collected are in English language. 
\par

A significant portion of the transcripts was sourced from Oyez \cite{oyez_website}, a multimedia judicial archive that provides publicly accessible Supreme Court transcripts. Additionally, courtroom interactions were extracted from legal television shows such as Paternity Court \cite{paternitycourt2013}, Support Court with Judge Vonda B. \cite{supportcourt2018}, and The People's Court \cite{peoplescourt2014}. While these shows are staged, they feature judges and legal professionals, and the dialogue mirrors courtroom conversations.
To preserve the integrity and legal accuracy of the dataset, careful verification was conducted to ensure that the transcripts adhered to standard legal frameworks and courtroom protocols.

\par
In total, the dataset comprises 1063 conversations featuring interactions between plaintiffs, defendants, lawyers, and judges. We also placed a special emphasis on collecting long conversations and the majority of the dialogues in the dataset average approximately 1000 words as shown in  \hyperref[fig:graphs]{Figure~\ref{fig:graphs}}. The distribution of manipulative and non-manipulative dialogues is given in \hyperref[tab:dataset]{Table~\ref{tab:dataset}}. Refer \hyperref[fig:bar]{Figure~\ref{fig:bar}} and \hyperref[fig:graphs]{Figure~\ref{fig:graphs}} for detailed visualization of the dataset.
To eliminate potential biases and standardize the dialogues, the original speakers’ names were replaced with their generic roles: “Plaintiff”, “Defendant”, “Plaintiff's Lawyer”, “Defendant's Lawyer”, and “Judge”.

\subsection{Labeling Schema and Annotation}
Building on insights from \cite{aldridge2007linguistic} and \cite{kadoch2000seduced}, we developed a multi-level labeling schema constituting three key components: (1) detecting manipulation, (2) identifying the primary manipulator, and (3) categorizing the specific manipulative techniques employed. The labeling schema and definitions are mentioned in the \hyperref[sec:appendix]{Appendix~\ref{sec:appendix}}.
\par
The three questions used for labeling and evaluation have been kept consistent throughout our research:
\begin{itemize}[noitemsep, topsep=0pt]
    \item Q1: Is the given dialogue manipulative?
    \item Q2: If yes, identify the primary manipulator.
    \item Q3: If the dialogue is manipulative, identify the manipulative techniques employed by the primary manipulator.
\end{itemize}
\begin{table}[htbp]  
    \centering
    \begin{tabular}{ccc}  
        \toprule
        \textbf{Dataset} & \textbf{Manipulative} & \textbf{Non-Manipulative} \\
        \midrule
        LegalCon & 663 & 400 \\
        \bottomrule
    \end{tabular}
    \caption{LegalCon Dataset Distribution}
    \label{tab:dataset}
\end{table}

\par
Through an extensive review of courtroom dialogues and existing studies on psychological manipulation in legal as well as other domains, we identified 11 frequently used manipulative techniques \cite{fischer2022then} \cite{mcdowell1991lawyer} \cite{aguado2015psychological}. We then consulted with psychology professors to verify and refine these categorizations. This schema is illustrated in appendix.

\par
While we initially explored using NLP techniques and LLMs for annotation, their performance proved inadequate. Hence, the four of us manually annotated the dataset, leveraging evidences and inferences from LLMs and prior research. 
To evaluate the reliability of the annotation process, we conducted a post-hoc inter-annotator agreement study on a subset of 100 dialogues. The annotators independently labeled this subset, and agreement was measured across the three tasks. Cohen's Kappa for Q1 and Q2 was 0.68 and 0.59 respectively.
For Q3, which is a multi-label classification task, we used Krippendorff’s Alpha, which yielded a score of 0.41, also indicating moderate agreement. 
As Q2 and Q3 were only applied when Q1 was marked as manipulative, the number of annotated items was filtered accordingly. These scores proved to be  consistent with expectations for subjective annotation tasks in legal and psychological domains. 
\par
Since by its inherent nature, manipulation is subjective and manipulation in a legal context especially so, we made an effort to only include the data points that had majority consensus in LegalCon.

\begin{figure*}[t]
    \centering
    \includegraphics[width=\textwidth]{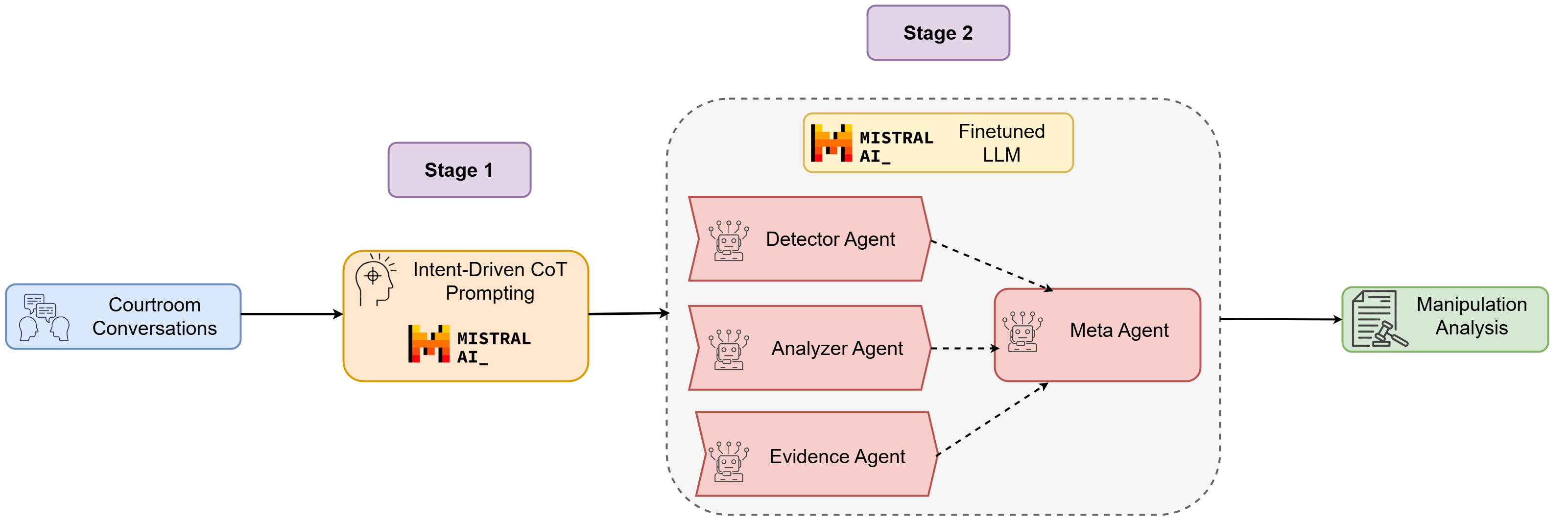}
    \caption{Overview of CLAIM: A two-stage framework for manipulation analysis}
    \label{fig:framework}
\end{figure*}

\section{Methodology}

In courtroom conversations, manipulation is often quiet and deeply rooted in the speaker's intent, rhetorical strategy, and power dynamics. Psychological studies show that individuals with a more substantial Theory of Mind (ToM) are better at interpreting others’ intentions and withstanding manipulation \cite{chen2024tombench}. A recent study also suggests that LLMs can improve their ToM performance when guided by structured reasoning techniques like Chain-of-Thought (CoT) prompting \cite{wei2022chain}.

Building on these studies, we propose \textbf{CLAIM}, a two-stage framework to improve manipulation analysis in courtroom dialogues: (1) Intent-Driven Chain-of-Thought Prompting to extract speaker intent and (2) a multi-agent decision framework to analyze manipulation using these inferred intents with the help of agents. This framework allows our method to reason with the speaker’s intent based on contextual evidence.
\hyperref[fig:framework]{Figure~\ref{fig:framework}} provides an overview of our framework, illustrating how each stage contributes to the final result.

\subsection{Intent-Driven CoT Prompting} 
In the first stage of our framework, we aim to uncover the underlying intent of each speaker. To achieve this, we implement Intent-Driven Chain-of-Thought (CoT) Prompting, which infers each speaker’s intents throughout the courtroom dialogue. This stage is inspired by the approach presented in the study \cite{ma2024detecting}. The resulting intent summaries provide a structured representation of speaker intent and act as intermediate reasoning scaffolds for subsequent analysis.

\begin{table*}[h!]
    \centering
    \small
    \begin{tabular}{lcccccccc}
        \toprule
        \multirow{2}{*}{\textbf{Experiment Setting}} 
        & \multicolumn{4}{c}{\textbf{Llama-3.1 8B}}  
        & \multicolumn{4}{c}{\textbf{Mistral 7B}} \\
        \cmidrule(lr){2-5} \cmidrule(lr){6-9}
        & $P$ & $R$ & ACC & $F_1$ & $P$ & $R$ & ACC & $F_1$ \\
        \midrule
        Zero-shot prompting
        & .713 & .600 & .609 & .614 
        & .618 & \textbf{.937} & .609 & .518 \\
        Few-shot prompting 
         & .653 & .811 & .622 & .598   
        & .717 & .747  & .667 & .664 \\
        CLAIM Stage 1 
         & - & - & - & -   
        & .753 & .674  & .667 & .670 \\
        \textbf{CLAIM (Our Work)} 
        & - & - & - & -  
        & \textbf{.757} & .821  & \textbf{.731} & \textbf{.727} \\
        \bottomrule
    \end{tabular}
    \caption{Results of the manipulation detection task on \textsc{LegalCon}. $P$, $R$, ACC, and $F_1$  stand for binary precision, binary recall, accuracy, and $F_1$ -score, respectively.}
    \label{tab:detection}
\end{table*}

\begin{table*}[h!]
    \centering
    \small
    \begin{tabular}{lcccccccc}
        \toprule
        \multirow{2}{*}{\textbf{Experiment Setting}} 
        & \multicolumn{4}{c}{\textbf{Llama-3.1 8B}}  
        & \multicolumn{4}{c}{\textbf{Mistral 7B}} \\
        \cmidrule(lr){2-5} \cmidrule(lr){6-9}
        & $P$ & $R$ & ACC & $F_1$ & $P$ & $R$ & ACC & $F_1$ \\
        \midrule
        Zero-shot prompting
        & .476 & .481 & .481 & .467 
        & .512 & .340 & .340 & .340 \\
        Few-shot prompting 
        & .484 & .449 & .449 & .454  
        & .557 & .481 & .481 & .489 \\
        CLAIM Stage 1 
         & - & - & - & -   
        & .419 & .526  & .526 & .464 \\
        \textbf{CLAIM (Our Work)} 
        & - & - & - & -  
        & \textbf{.608} & \textbf{.609} & \textbf{.609} & \textbf{.602}\\
        \bottomrule
    \end{tabular}
    \caption{Results of the primary manipulator identification task \textsc{LegalCon}. $P$, $R$, ACC, and $F_1$  stand for binary precision, binary recall, accuracy, and $F_1$ -score, respectively.}
    \label{tab:pm}
\end{table*}

\begin{table*}[h!]
    \centering
    \small
    \begin{tabular}{lcccccccccc}
        \toprule
        \multirow{2}{*}{\textbf{Experiment Setting}} 
        & \multicolumn{5}{c}{\textbf{Llama-3.1 8B}}  
        & \multicolumn{5}{c}{\textbf{Mistral 7B}} \\
        \cmidrule(lr){2-6} \cmidrule(lr){7-11}
        & $P$ & $R$ & ACC & $F_1$ & $Jc$ & $P$ & $R$ & ACC & $F_1$ & $Jc$\\
        \midrule
        Zero-shot prompting
        & .1899 & .3379 & .2436 & .2082 & .3106 
        & .1715 & .3988 & .0385 & .1387 & .1265 \\
        Few-shot prompting 
        & .1515 & .4198 & .1346 & .1915 & .2271 
        & .2392 & .3394 & .2179 & .2118 & .3145 \\
        CLAIM Stage 1 
         & - & - & - & - & -  
        & .2582 & .4201  & .2115 & \textbf{.2452} & .3028\\
        \textbf{CLAIM (Our Work)}
        & - & - & - & - & -
        & \textbf{.2639} & \textbf{.4300} & \textbf{.2564} & .2354 & \textbf{.3618} \\
        \bottomrule
    \end{tabular}
    \caption{Results of the manipulation technique identification task on \textsc{LegalCon}. $P$, $R$, ACC, $F_1$ and $Jc$ stand for binary precision, binary recall, accuracy, $F_1$-score, and Jaccard coefficient, respectively.}
    \label{tab:tactics}
\end{table*}

\subsection{Multi-Agent Framework}

Manipulation analysis in long-form courtroom conversations is a complex task that requires context-aware decision-making. Relying on a single model to manage the full complexity of such discourse often leads to brittle outputs and poor interpretability. To address this, the second stage of our framework adopts a multi-agent architecture, where each agent is responsible for a specific subtask within the manipulation analysis pipeline.
LLM agents are particularly effective for such multi-step decision-making processes, as they support decomposition of reasoning, evidence aggregation, and inter-agent communication. In our framework, each agent operates independently but shares intermediate outputs with other agents to collaboratively arrive at a final judgment. 
\par
To optimize the performance of our agents for legal-domain reasoning, we fine-tuned the Mistral-7B \cite{jiang2023diego} language model on LegalCon, a curated dataset of courtroom and legal exchanges explained in Section~\ref{sec:Constructing LegalCon}. We use QLoRA \cite{dettmers2023qlora}, a memory-efficient parameter-efficient fine-tuning \cite{xu2023parameter} method that enables low-resource adaptation of large models. This framework results in lightweight, high-performance agents optimized for courtroom manipulation analysis, organized into four specialized components, each responsible for a specific subtask:
\begin{itemize}[noitemsep, topsep=0pt]
    \item \textbf{Detector Agent:} This agent is designed to determine whether the courtroom dialogues contain manipulation.
    \item \textbf{Analyzer Agent:} This agent is designed to identify the primary manipulator and classify the manipulation techniques used by the primary manipulator in the dialogue.
    \item \textbf{Evidence Agent:} This agent is designed to extract evidence from the dialogue that substantiates the manipulator and techniques used by them.
    \item \textbf{Meta Agent:} This agent is designed to aggregate the outputs from all agents, generating a final set of labels.
\end{itemize}

Agents receive the courtroom dialogue along with the intents of each speaker generated in Stage 1 as input. These intent representations provide additional reasoning context, allowing agents to align manipulation judgments with inferred speaker goals. Then the meta agent summarizes and compiles the final result.
This framework enables more robust reasoning and improves interpretability, as each decision is traceable to an agent's role and output.

\section{Experiments}
\subsection{Experimental Settings}
We conducted experiments on three tasks using the LegalCon dataset to assess the performance of CLAIM and SoTA models in analyzing manipulation in courtroom dialogues. These tasks include: Manipulation Detection, Primary Manipulator Classification, and Manipulation Technique Classification.
For the experimental data, we randomly split the dataset into 70\% for training, 15\% for validation, and 15\% for testing. We compared two models, Mistral-7B \cite{jiang2023diego} and Llama 3.1 8B \cite{grattafiori2024llama}, across four experimental settings: zero-shot prompting, few-shot prompting, CLAIM Stage 1 alone and CLAIM.

In the zero-shot prompting, courtroom dialogues were presented directly to the models with instructions to detect whether manipulation occurred. In the few-shot prompting, we provided each model with two non-manipulative and three manipulative courtroom conversations as in-context examples along with the task prompt. The format for both zero-shot and few-shot prompting are outlined in the Appendix. Additionally, we experimented with CLAIM Stage 1 alone, where speaker intents were inferred from the dialogues using CoT prompting. Manipulation analysis was then performed based solely on these inferred intents, and corresponding results were calculated.
In our proposed framework CLAIM, we applied our two-stage Intent-Driven with a Multi-Agent framework, where the inferred intents of speakers guided specialized agents to analyze manipulation. The agents were powered by a Mistral-7B model fine-tuned using QLoRA,  a memory-efficient PEFT method. The fine-tuning was performed with a learning rate of 1e-4 to optimize the model for legal-domain reasoning and manipulation detection.
All experiments were conducted on an MSI GeForce RTX 3060 GPU. Both models were tested at temperatures of 0.4 and 0.6, and the models performed most consistently and accurately at a temperature of 0.4. 

\subsection{Experimental Results}
\hyperref[tab:detection]{Table~\ref{tab:detection}} presents the experimental results for manipulation detection, comparing CLAIM against baseline models using zero-shot and few-shot prompting as well as with CLAIM Stage 1. The results indicate that CLAIM outperforms the baseline models, achieving higher accuracy. This demonstrates the effectiveness of our framework in detecting manipulation more reliably than traditional prompting techniques.
\hyperref[tab:pm]{Table~\ref{tab:pm}} presents the results for primary manipulator identification, a challenging and inherently subjective task. The findings indicate that all models face difficulties in accurately identifying the manipulator. Further analysis reveals that this challenge arises from the models frequently misattributing manipulative intent. Despite this, CLAIM demonstrates a notable improvement over baseline methods.
\hyperref[tab:tactics]{Table~\ref{tab:tactics}} presents the results for identifying manipulation techniques employed by the primary manipulator. The models do not perform well and exhibit relatively low accuracy. Since this is a multi-label classification task, correctly identifying all the techniques is challenging. Traditional accuracy metrics may not fully capture performance. To address this, we used Jaccard Similarity Coefficient, which is used to calculate the overlap between predicted and actual manipulation techniques as shown in equation (1).

\begin{equation}
J(A, B) = \frac{|A \cap B|}{|A \cup B|}
\end{equation}

where \( A \) is the set of true manipulation techniques for a given instance, and \( B \) is the set of predicted techniques.

CLAIM achieved the highest Jaccard score of 0.3618. However, due to the subjective nature of manipulation detection, distinguishing certain techniques remains challenging and open to debate.

\section{Conclusion and Future Work}
This study introduces LegalCon, a dataset of courtroom conversations aimed at detecting and analyzing manipulation. Alongside this, we propose CLAIM, a two-stage Intent-driven Multi-Agent framework, to enhance the detection and analysis of manipulation in courtroom conversations.
Extensive experiments showed that our method consistently outperformed baseline models on various prompting techniques. 
However, the models struggled to accurately identify specific manipulative techniques, revealing a critical limitation. These findings highlight the inherently subjective and nuanced nature of manipulation.
With LegalCon and CLAIM we hope to address a critical gap in NLP research at the intersection of law and manipulative language. 
Since legal decisions are lasting and influential we hope that this lays a necessary foundation for further work in this field.
\par
Future work could focus on expanding the LegalCon dataset to include different types of cases and create a more comprehensive dataset. Multi-lingual transcripts can be incorporated to enhance diversity and enable cross-cultural analysis of manipulative language.
Multi-modal frameworks can be explored to yield deeper insights into manipulation dynamics. 
Integrating these advancements into real-world legal settings is a valuable opportunity and moving forward, efforts should focus on the responsible deployment of such models to support legal professionals and promote fairness in courtrooms.
Given that manipulation detection in the legal domain is a relatively underexplored area, further research in this field could provide valuable insights and open up new avenues for improving legal processes and contributing to the advancement of the application of technology in the legal domain.\\
\section{Limitations}
While our proposed framework demonstrates promising results, there are several limitations to consider:\\
\textbf{Subjectivity in Manipulation Detection}:
Manipulation, by nature, is subjective, which makes it challenging for models to accurately identify manipulative behavior. Since there are no well-defined standard limits distinguishing between manipulative and non-manipulative behaviors, especially in arguments and debates, it remains particularly complex.\\
\textbf{Dataset Annotation Challenges}:
Despite our efforts to ensure high-quality annotations, labeling manipulation, especially for specific techniques remains subjective. While the annotators made efforts to minimize bias, human interpretation is influenced by personal perspectives. This subjectivity in labeling may affect the consistency and reliability of the dataset, which in turn could impact the model's training and overall performance.\\
\textbf{Limited Generalizability of the LegalCon Dataset}:
LegalCon dataset is limited in scope, covering only a specific set of case types. Hence, the model may not generalize well to other legal contexts or jurisdictions.\\
\textbf{Limited Generalizability of CLAIM framework}:
The CLAIM framework was developed for the LegalCon dataset, and optimized particularly to address challenges posed by longer courtroom conversations. However, it may struggle to generalize to shorter dialogues outside of courtroom settings. Additionally, the framework's complexity might be excessive for such tasks, making it less suitable for simpler or more informal interactions.\\
\textbf{Computational Constraints}: Fine-tuning LLMs requires significant computational resources but due to hardware limitations, we were restricted in terms of batch size and the number of training epochs. The fine-tuning process was conducted on a single MSI GeForce RTX 3060, which limited our ability to experiment with larger models.
 \section{Ethics Statement}

 All data used in the LegalCon dataset was sourced from publicly available transcripts, including court proceedings and legally staged courtroom television shows. We ensured that no personally identifiable information (PII) was retained. Speaker names were anonymized and replaced with generic role labels such as "Plaintiff", "Defendant", "Plaintiff's Lawyer", "Defendant's Lawyer" and "Judge" to protect identities and maintain legal neutrality.
\bibliographystyle{acl_natbib}

\vspace{\baselineskip}
\appendix

\section{Appendix}
\subsection{Labeling Schema for LegalCon}
\begin{figure*}[htbp] 
    \centering
    \includegraphics[width=\textwidth]{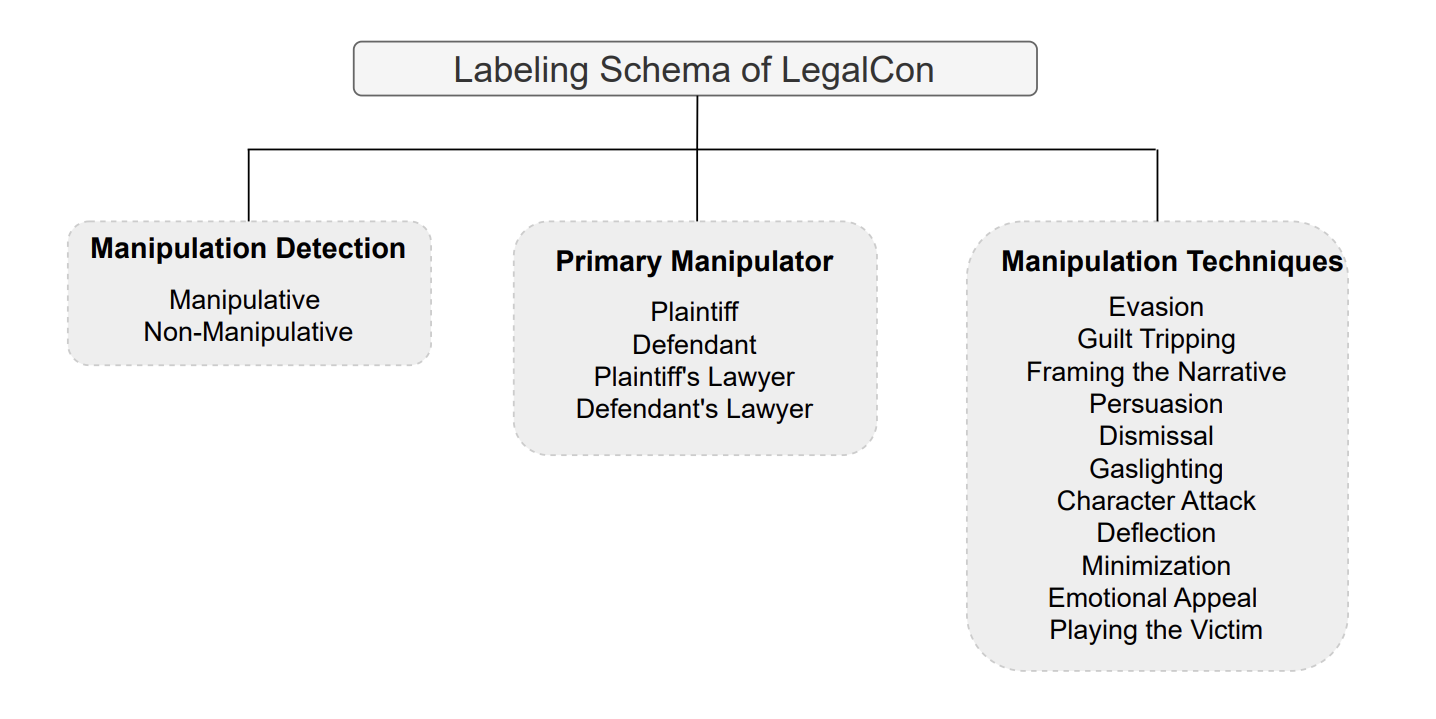}
    \caption{Labeling schema showing 11 manipulative techniques used for annotation in LegalCon.}
    \label{fig:schema}
\end{figure*}

Definitions of the 11 manipulative techniques selected and used for labeling LegalCon listed in \hyperref[fig:schema]{Figure~\ref{fig:schema}} are:\\
\begin{enumerate}
    \item \textbf{Gaslighting}: A form of psychological manipulation where a person makes someone doubt their perceptions or sanity by denying the truth or altering reality.
    \item \textbf{Guilt tripping}: A manipulative tactic where someone tries to make another feel guilty to control their behavior. It often involves exaggerating the impact of their actions or making them feel responsible for things not their fault.
    \item \textbf{Persuasion}: Influencing someone's beliefs or actions through reasoning or appealing to their interests.
    \item \textbf{Evasion}: The act of avoiding a question, responsibility, or engagement, while manipulation involves influencing or controlling someone or something unfairly to one's advantage.
    \item \textbf{Framing the narrative}: Selectively highlighting certain aspects of a story to influence an audience's perception and understanding.
    \item \textbf{Dismissal}: Ignoring other people’s concerns or questions with the aim to monopolize information and control other people’s choices and decisions.
    \item \textbf{Character Attack}: Deliberate and sustained effort to damage a person's reputation, often through manipulation.
    \item \textbf{Deflection}: Avoiding addressing true feelings or actions by shifting focus onto someone or something else. Deflection may also be used to evade responsibility or to place blame on others, thereby avoiding accountability.
    \item \textbf{Minimization}: Downplaying or trivializing events, emotions, or experiences to reduce their perceived importance. Often used to invalidate feelings or diminish the impact of harmful behavior.
    \item \textbf{Emotional appeal}: Attempting to influence others by exploiting emotions instead of using logic or evidence. Often relies on misleading or sentimental language to provoke fear, guilt, or sympathy and bypass rational judgment.
    \item \textbf{Playing the victim}: Exaggerating or fabricating an event, experience, or emotion to portray themselves as a victim in the situation when in reality they are not a victim.
\end{enumerate}
\onecolumn

\label{sec:appendix}

\subsection{Example Results of CLAIM framework}
\begin{tcolorbox}[colback=white, colframe= orange!80!black, title= \textbf{CLAIM Stage 1 Results}]
\ttfamily
\textbf{Plaintiff's Intent}:
The Plaintiff is claiming that she was not paid for her services, was provided uncertified training, and cannot get her training certificate. She is suing the 
Defendant,  for every penny of the \$3,375 she is owed.

\textbf{Defendant's Intent}:
The Defendant is stating that she has a counterclaim for \$5,000 for breach of contract, as the Plaintiff, did not complete the required 200 hours of training and left a hole in her schedule, owing \$1,040.

\end{tcolorbox}

\begin{tcolorbox}[colback=white, colframe=purple!40!black, title= \textbf{CLAIM Stage 2 Results}]
\ttfamily
\textbf{Detector Agent Result:}\\
Yes\\
\textbf{Analyzer Agent Results:}\\
Primary Manipulator -  Plaintiff\\
Manipulative Techniques -  Emotional appeal, Framing the narrative, Character attack\\
\textbf{Evidence Agent Results:}\\
Primary Manipulator - Plaintiff\\
Manipulative Techniques - Emotional appeal, Framing the narrative, Character attack\\
\textbf{Meta Agent Results:}\\
Manipulation Present - Yes\\
Primary Manipulator -  Plaintiff\\
Techniques Used - Emotional appeal, Framing the narrative, Character attack\\
\end{tcolorbox}

\subsection{Prompt Template in Experiments}

\begin{tcolorbox}[colback=white, colframe=  green!50!black, title=\textbf{Zero-shot Prompting}]
\ttfamily
You are a manipulation detection expert. Your task is to analyze this dialogue and determine if there is manipulation present.

\textbf{\#\#\# Instructions:}
\begin{enumerate}[leftmargin=*]
  \item Determine whether any manipulation is present. Answer with only ‘Yes’ or ‘No’. Do not add anything else.
  \item If manipulation is present, then identify the primary manipulator. Answer using only the speaker label. Don't add anything else.
  \item Identify the manipulation techniques used by them, selecting only from the allowed list. Provide the techniques using only those exact terms. Do not add anything else.
\end{enumerate}

<insert dialogue>
\end{tcolorbox}

\begin{tcolorbox}[colback=white, colframe= red!60!black, title=\textbf{Few-shot Prompting}]
\ttfamily
You are a manipulation detection expert. Your task is to analyze this dialogue and determine if there is manipulation present. Here are five examples:

\noindent{Example 1:}\\
<insert manipulative\_dialogue1>\\
<insert manipulative\_answer1>

\noindent{Example 2:}\\
<insert manipulative\_dialogue2>\\
<insert manipulative\_answer2>

\noindent{Example 3:}\\
<insert manipulative\_dialogue3>\\
<insert manipulative\_answer3>

\noindent{Example 4:}\\
<insert nonmanipulative\_dialogue1>\\
<insert nonmanipulative\_answer1>

\noindent{Example 5:}\\
<insert nonmanipulative\_dialogue2>\\
<insert nonmanipulative\_answer2>

\textbf{\#\#\# Instructions:}
\begin{enumerate}[leftmargin=*]
  \item Determine whether any manipulation is present. Answer with only ‘Yes’ or ‘No’. Do not add anything else.
  \item If manipulation is present then identify the primary manipulator. Answer using only the speaker label. Don't add anything else.
  \item Identify the manipulation techniques used by them, selecting only from the allowed list. Provide the techniques using only those exact terms. Do not add anything else.
\end{enumerate}

<insert dialogue>
\end{tcolorbox}

\vspace{2em}
\begin{tcolorbox}[colback=white, colframe=yellow!60!black,title=\textbf{CLAIM Stage 1 Prompt}]
\ttfamily
You are reading a transcript from a courtroom conversation.
\begin{enumerate}[leftmargin=*]
    \item Carefully read the dialogue.
    \item Think step-by-step about what the plaintiff's and defendant's statements suggest.
    \item Reason about the plaintiff's and defendant's goals or motives behind their words.
    \item Summarize the plaintiff's and defendant's intent in a sentence.
\end{enumerate}
\end{tcolorbox}

\vspace{2em}

\begin{tcolorbox}[colback=white, colframe=blue!50!black, title=\textbf{CLAIM Stage 2 Prompts}]
\ttfamily

\ttfamily
\textbf{Detector Agent}:\\
You are a manipulation detection expert. Your task is to analyze the dialogue and the corresponding intents to determine whether manipulation is present.

\textbf{\#\#\# Instructions:}
\begin{enumerate}[leftmargin=*]
  \item Read the dialogue carefully.
  \item Analyze it in the context of the provided intents.
  \item Determine whether any manipulation is present. Answer with only ‘Yes’ or ‘No’. Do not add anything else.
\end{enumerate}

\textbf{Analyzer Agent}:\\
You are responsible for identifying manipulation analysis within a courtroom dialogue using both the dialogue and the inferred speaker intents.

\textbf{\#\#\# Instructions:}
\begin{enumerate}[leftmargin=*]
  \item Identify the primary manipulator. Answer using only the speaker label. Do not add anything else.
  \item Identify the manipulation techniques used by them, selecting only from the allowed list. Provide the techniques using only those exact terms. Do not add anything else.
 
\end{enumerate}

\textbf{Evidence Agent}:\\
You are tasked with validating the manipulation analysis based on the dialogue.

\textbf{\#\#\# Instructions:}
\begin{enumerate}[leftmargin=*]
  \item Review whether the identified primary manipulator and manipulative techniques are correct.
  \item If incorrect, update them. Answer only with the updated result.

\end{enumerate}

\end{tcolorbox}

\end{document}